\documentclass[letterpaper, 10 pt, conference]{ieeeconf}  

\IEEEoverridecommandlockouts                              

\usepackage{graphics} 
\usepackage{epsfig} 
\usepackage{times} 
\usepackage{amsmath,amssymb,amsfonts}
\usepackage{multirow}
\usepackage{tabularx}
\usepackage{balance}

\usepackage{calc}

\usepackage{caption}
\DeclareCaptionLabelSeparator{dottilde}{.~~}
\usepackage{hyperref}
\usepackage{cleveref}
\usepackage{booktabs}

\usepackage{algorithm}
\usepackage{algpseudocode}
\usepackage{amsmath}

\usepackage{booktabs}
\usepackage{makecell}  
\usepackage{graphicx}  
\usepackage{color} 
\usepackage{soul} 

\usepackage{amssymb}

\title{\LARGE \bf
Adap-RPF: Adaptive Trajectory Sampling for Robot Person Following in Dynamic Crowded Environments
}

\author{
Weixi Situ\textsuperscript{*}$^{1}$, 
Hanjing Ye\textsuperscript{*}$^{1}$, 
Jianwei Peng$^{2}$, 
Yu Zhan$^{1}$, 
and Hong Zhang\textsuperscript{\dag, 1}\textit{, Life Fellow, IEEE}
\thanks{\textsuperscript{+}Equal contributions.}
\thanks{$^{1}$W. Situ, H. Ye, Y. Zhan and H. Zhang are with the Robotics and Computer Vision (RCV) Laboratory, Southern University of Science and Technology (SUSTech), Shenzhen, China.}
\thanks{$^{2}$J. Peng is with SUSTech.}
\thanks{$\dag$ Corresponding author: Hong Zhang (hzhang@sustech.edu.cn)}
}

\begin{document}

\maketitle
\thispagestyle{empty}
\pagestyle{empty}

\begin{abstract}
Robot person following (RPF) is a core capability in human–robot interaction, enabling robots to assist users in daily activities, collaborative work, and other service scenarios. However, achieving practical RPF remains challenging due to frequent occlusions, particularly in dynamic and crowded environments. Existing approaches often rely on fixed-point following or sparse candidate-point selection with oversimplified heuristics, which cannot adequately handle complex occlusions caused by moving obstacles such as pedestrians. To address these limitations, we propose an adaptive trajectory sampling method that generates dense candidate points within socially aware zones and evaluates them using a multi-objective cost function. Based on the optimal point, a person-following trajectory is estimated relative to the predicted motion of the target. We further design a prediction-aware model predictive path integral (MPPI) controller that simultaneously tracks this trajectory and proactively avoids collisions using predicted pedestrian motions. Extensive experiments show that our method outperforms state-of-the-art baselines in smoothness, safety, robustness, and human comfort, with its effectiveness further demonstrated on a mobile robot in real-world scenarios.
\end{abstract}
\section{INTRODUCTION}
Robot person following (RPF) is a fundamental capability in human–robot interaction (HRI), enabling a wide range of applications \cite{islam2019person, goodrich2008human, eirale2025human}. However, existing methods based on fixed-point following or sparse candidate selection with simplified heuristics struggle to cope with frequent occlusions of the target by static objects (e.g., walls or furniture) or dynamic obstacles (e.g., other pedestrians), posing significant challenges for deploying RPF in real-world environments.

Conventional RPF methods adopt a fixed-point following strategy, tracking a target point that maintains a constant relative distance and orientation to the target. The relative orientation is typically behind \cite{leigh2015person, peng2024dual, zhao2024human}, ahead \cite{nikdel2018hands, nikdel2021lbgp}, or side by side \cite{morales2012people, peng2023mpc}, under the assumption that the target remains continuously visible. In complex and crowded environments, this assumption often breaks down, as rigidly maintaining a fixed relative position can cause the robot to lose sight of the target person when nearby people or objects temporarily obstruct the view, or render the target point unreachable when it is occupied, leading to frequent occlusions and tracking failures. To address occlusions, some methods \cite{ye2025rpf,lee2018icra} predict potential search regions to reacquire the target person. However, these reactive strategies can induce oscillatory behavior when the search space is constrained by surrounding pedestrians.
\begin{figure}[t]
    \centering
    \includegraphics[width=\linewidth]{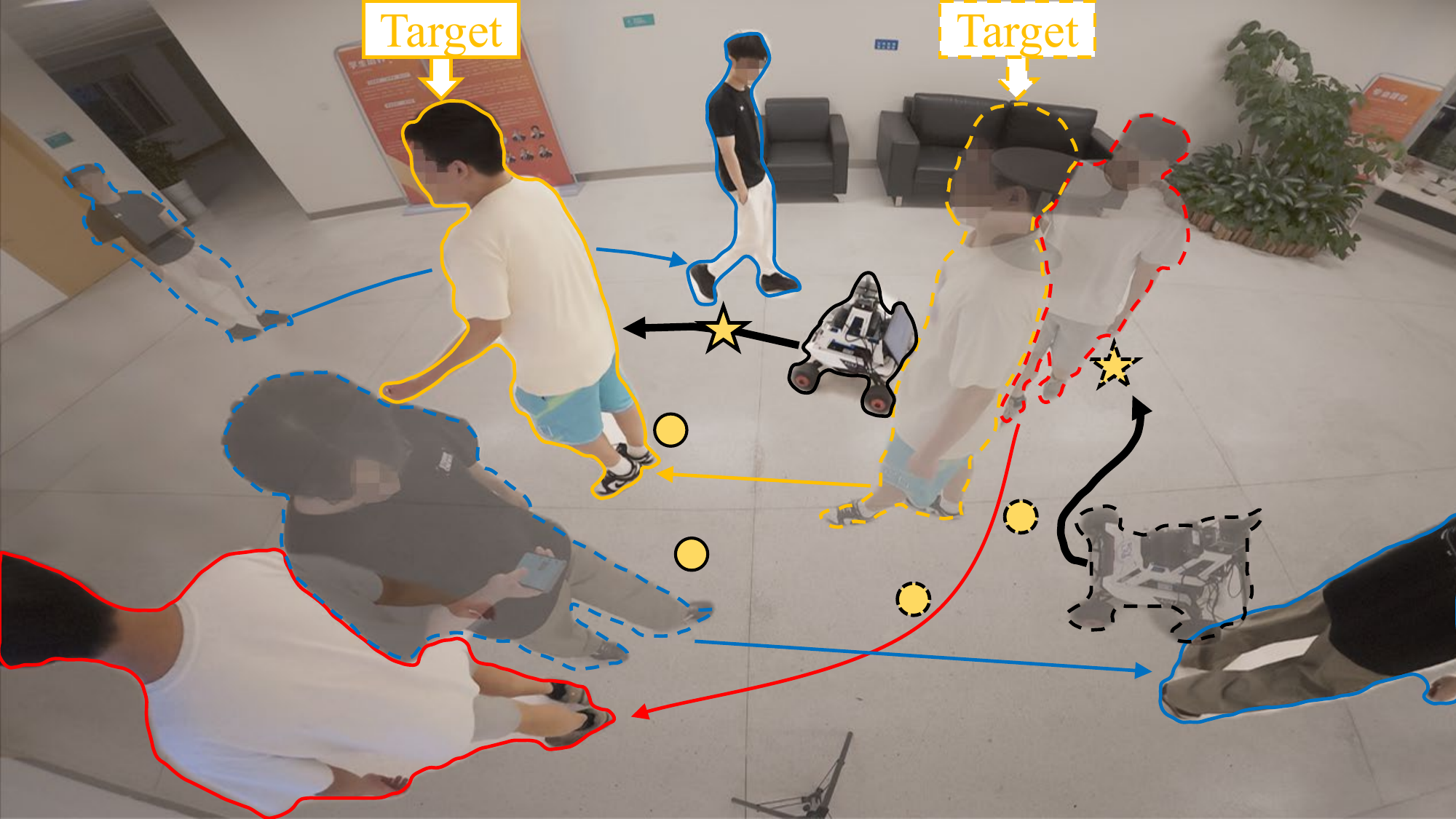}
    \caption{\textbf{Adap-RPF in a real-world dynamic crowded environment.} The robot proactively avoids dynamic occlusions (the pedestrian outlined in red) by adaptively sampling trajectory based on predicted human motion (colored arrows). Dashed outlines indicate agent positions at time $T_0$, while solid outlines indicate their positions at time $T_1$. The golden star marks the selected optimal following point, and golden circles indicate the candidate points.}
    \label{fig:figure1}
    \vspace*{-0.25in}
\end{figure}

An alternative is proactive planning to prevent occlusion and potential collisions. A common strategy \cite{hoeller2007accompanying, yao2021laser} involves: i) pre-defining several candidate following points around the target, and ii) dynamically switching among them based on the situation. These approaches establish a small set of candidate points or regions around the target and dynamically switch among them according to heuristic rules, enabling proactive planning to avoid obstacles and potential collisions while maintaining continuous tracking. Although these methods exemplify the general \textit{sample–evaluate–plan} paradigm, they tend to oversimplify both candidate point selection and evaluation, often assuming back-following as the safest option and neglecting dynamic occlusions. Consequently, sparse sampling with simplified heuristics limits their effectiveness in dynamic, crowded environments where occlusions and collisions are prevalent.

Recently, several methods \cite{repiso2017line,Leisiazar2025ahead,song2023safe, wang2024continuous} have been proposed to directly estimate the next robot action through a unified model, based on either deep-learning \cite{Leisiazar2025ahead, leisiazar2023mcts} or optimization \cite{repiso2017line,song2023safe,wang2024continuous}. Deep-learning-based approaches often suffer from poor generalization and limited interpretability, reducing their effectiveness in unseen scenarios. In contrast, optimization-based methods formulate the problem by integrating multiple constraints, such as target visibility and collision avoidance, into a unified optimization framework \cite{wang2024continuous}. However, this often sacrifices proxemic comfort, leading to overly distant following behaviors. Moreover, their formulation only leverages the predicted target trajectory for visibility optimization, while neglecting the predicted motions of surrounding pedestrians for proactive collision avoidance, thereby limiting adaptability in dynamic environments.

To address these limitations, we propose \textbf{Adap-RPF}, a hierarchical planning framework that proactively integrates target visibility, collision avoidance, and human comfort. Unlike fixed-point following or sparse candidate-point selection methods \cite{hoeller2007accompanying,yao2021laser}, our approach performs target-centric adaptive trajectory sampling over a dense set of candidate points generated along the predicted target
motion within a semi-annular region respecting the target’s personal and social zones. Candidates are evaluated through a multi-objective cost function considering both prediction-aware factors (e.g., occlusion, proximity) and observation-based factors (e.g., distance, travel, stickiness). The optimal point is then extrapolated into a following trajectory, conditioned on the target’s predicted motion.
Building on this trajectory, we further develop a prediction-aware model predictive path integral (MPPI) controller that tracks the reference trajectory while proactively avoiding collisions based on predicted pedestrian motions. This predictive integration into an MPC-style controller is inspired by related work in crowd navigation~\cite{le2024social,poddar2023crowd}, but represents the first attempt in the context of RPF.
Experimental results validate that our framework is both practical and efficient, enabling smoother and safer person-following behaviors compared with state-of-the-art baselines, even in highly dynamic environments. Moreover, we demonstrate the real-world applicability of our method through experiments on a mobile robot (Fig.~\ref{fig:figure1}).

\section{RELATED WORKS}
This section presents a literature review on existing RPF approaches, with particular attention to their proactive strategies for dealing with occlusion and their potential incorporation of collision avoidance functions.
\vspace{0.15\baselineskip}
\subsection{Robot Person Following with Multiple Predefined Points}
Some methods define multiple candidate following points around the target and switch among them based on heuristic rules. For example, Hoeller \textit{et al.} \cite{hoeller2007accompanying} discretize a semicircular region behind the target into five candidate points and prefer the point without collision risk, treating back-following as the preferred strategy. However, this method simplifies point sampling and evaluation, leaving the method ill-suited for crowded environments where dynamic occlusions and collision risks occur frequently. Yao \textit{et al.} \cite{yao2021laser} switches between rear and side positions based on static obstacle detection and corridor geometry, while Vu \textit{et al.} \cite{vu2025autonomous} models an elliptical region around the target’s motion and selects feasible points according to obstacle presence and distance. These methods address static occlusion based on static obstacle or environmental structure recognition, which is not generalized to dynamic environments where frequent and dynamic occlusions occur. 

To address these challenges, we employ a denser point-sampling strategy combined with a multi-objective evaluation framework that jointly accounts for proximity constraints~\cite{hall1963system}, target visibility, collision avoidance, and motion smoothness. The selected optimal point is then extended into a person-following trajectory by incorporating the predicted motion of the target. This trajectory-sampling scheme is both computationally efficient and adaptive to dynamic environments.

\vspace{0.15\baselineskip}
\subsection{Robot Person Following with A Unified Model}
Recently, several approaches have explored unified models that directly generate the robot's next action. Such methods can be broadly categorized into learning-based methods \cite{Leisiazar2025ahead, leisiazar2023mcts} and optimization-based methods \cite{repiso2017line,song2023safe,wang2024continuous}. Learning-based methods, such as \cite{Leisiazar2025ahead}, train reinforcement learning agents on randomized human trajectories with reward functions encoding preferred distance and orientation, thereby enabling adaptive action selection. However, these approaches require extensive training, generalize poorly, and offer limited interpretability, which restricts their effectiveness in unseen scenarios. 

In contrast, optimization-based approaches explicitly incorporate task constraints into the planning formulation. For example, Repiso \textit{et al.} \cite{repiso2017line} formulate multiple person-following costs within the anticipate kinodynamic planning framework, switching between rear and side positions in narrow passages. Yet, this approach relies on a global map and assumes the target’s global destination is known in advance, limiting its applicability in dynamic, unknown environments. Wang \textit{et al.} \cite{wang2024continuous} embed objectives such as target detectability, tracking accuracy, motion smoothness, and obstacle avoidance into a nonlinear MPC framework for multi-target following. Yet, their approach struggles to balance visibility with tracking and often overlooks proxemic comfort. Moreover, it generalizes poorly in highly dynamic settings, as it does not leverage future trajectories of surrounding pedestrians, leading to limited proactive collision avoidance. 

By contrast, our Adap-RPF framework explicitly incorporates predicted trajectories of all observed pedestrians into the controller, enabling proactive and reliable collision avoidance. Furthermore, rather than embedding all RPF-related objectives directly into the controller optimization, we delegate most objectives to the adaptive trajectory sampling module. This decomposition reduces the computational burden of the controller and allows faster response to dynamic pedestrians.
\section{METHODOLOGY}
\begin{figure}[t]
    \centering
    \includegraphics[width=\linewidth]{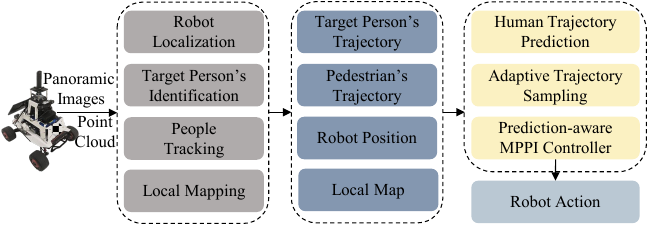}
        \caption{\textbf{RPF system pipline.} We integrate Adap-RPF framework into the system, where the light yellow modules indicate our contributions. Our framework consists of three components: human trajectory prediction, adaptive trajectory sampling and prediction-aware MPPI controller. Additional RPF system modules, are adopted from work \cite{ye2025rpf} and are not the focus of this paper. Overall, the proposed RPF system can locate, track, and follow a target person while proactively avoiding occlusions in dynamic, crowded environments.}
        \label{fig:system}
        \vspace*{-0.25in}      
\end{figure}
In this section, we present Adap-RPF, a hierarchical planning framework that leverages human trajectory prediction (Sec.~\ref{sec:prediction}) to proactively address target visibility, proxemic comfort, collision avoidance, and motion smoothness. The key innovation lies in the adaptive trajectory sampling module (Sec.~\ref{sec:samping}), which generates a dense set of candidate following points within the target-centric social space, rather than relying on fixed or sparsely predefined samples. These candidates are then evaluated against predicted pedestrian trajectories to anticipate dynamic occlusions and potential collisions, with the optimal point selected to form a person-following trajectory aligned to the target’s predicted motion. We subsequently design a prediction-aware MPPI controller to track this reference trajectory (Sec.~\ref{sec:mppi}), which incorporates predicted trajectories of surrounding pedestrians into its optimization process, thereby ensuring proactive and reliable collision avoidance. 

The complete RPF framework is illustrated in Fig.~\ref{fig:system}. Our approach tracks people and identifies the target person based on \cite{ye2024person, ye2023robot}. Additionally, we utilize FAST-LIO~\cite{xu2021fast} for robot localization and construct a local map by filtering the 3D LiDAR scan to retain points within a narrow band around the robot’s body height, which are then projected onto a 2D plane. Our Adap-RPF mainly utilizes the aforementioned perception information to proactively follow the target person.

\subsection{Human Trajectory Prediction}\label{sec:prediction}
Building on the people detection, tracking, and target identification modules, we obtain the global positions of the target and surrounding pedestrians. At each timestep, every pedestrian’s position is appended to a FIFO-updated queue. Their historical trajectories are then used to predict an $\mathrm{N}$-step future trajectory with a constant velocity model (CVM), which offers high computational efficiency. Notably, our framework is agnostic to the choice of prediction model, allowing seamless integration of alternative trajectory predictors.

\begin{figure*}[htbp]
  \centering
  \includegraphics[width=0.85\textwidth]{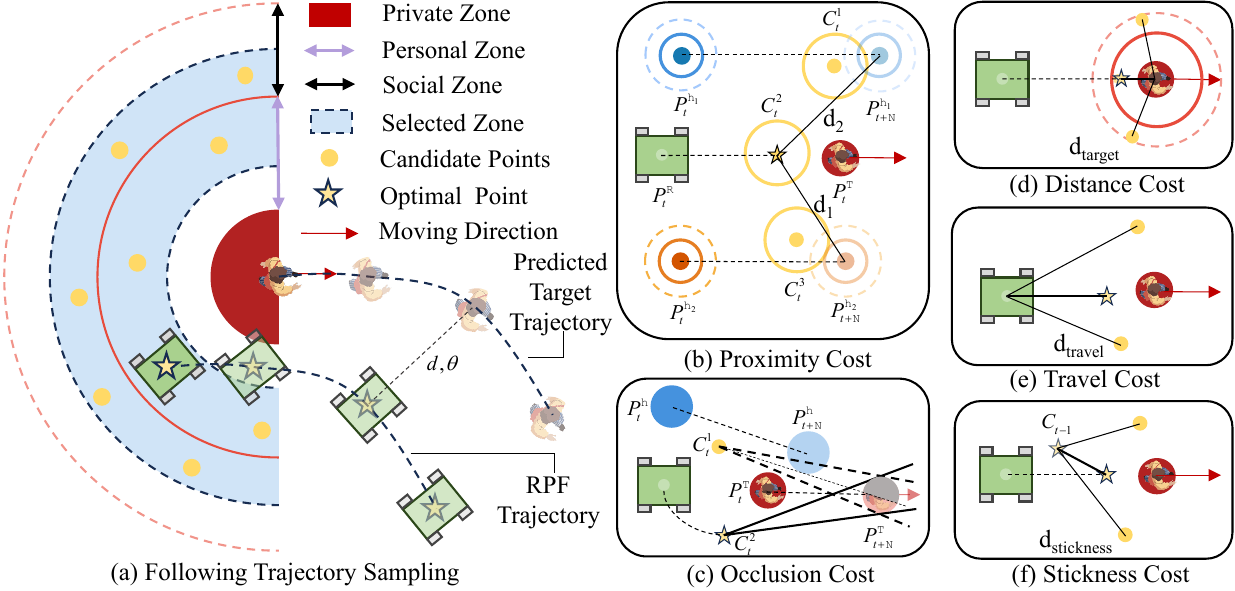} 
\caption{
\textbf{Target-centric Adaptive Following Trajectory Sampling.} 
(a) \textbf{Following Trajectory Sampling.} Candidate following points are generated using Sobol sampling within a target-centric semi-annular region defined by the target’s personal and social zones. The candidates are evaluated using a multi-objective cost function that accounts for target visibility, collision risk, social compliance, and smoothness. compliance, and smoothness. A following trajectory is then constructed relative to the predicted target trajectory using an offset ($d$, $\theta$) based on the selected optimal point.
(b) \textbf{Proximity Cost (Eq.~\ref{eq:proximity}).} Minimum distance $d_{1}$ to surrounding pedestrians is computed from predicted trajectories and social zones; candidates intruding into personal space (e.g., $\mathbf{C}_{t}^{3}$, $\mathbf{C}_{t}^{1}$) are discarded, and $\mathbf{C}_{t}^{2}$ is selected.
(c) \textbf{Occlusion Cost (Eq.~\ref{eq:occlusion}).} Occlusion is estimated via IoU between pedestrian  and target projections ($\mathbf{P}_{t+N}^{h}$ and $\mathbf{P}_{t+N}^\mathrm{Tar}$) using predicted trajectories; $\mathbf{C}_{t}^{1}$ is rejected due to partial occlusion (light gray), whereas $\mathbf{C}_{t}^{2}$ remains visible.
(d) \textbf{Distance Cost (Eq.~\ref{eq:dis}).} Penalizes deviation from the desired spacing to the target.
(e) \textbf{Travel Cost (Eq. ~\ref{eq:travel}).} Encourages minimal path effort from the robot’s current position (green rectangle $\mathbf{P}_{t}^{R}$).  
(f) \textbf{Stickiness Cost (Eq.~\ref{eq:stick}).}  Promotes continuity by favoring points near the previously selected following point.
} 
  \label{fig:sampling}
\end{figure*} 
\subsection{Adaptive Trajectory Sampling}\label{sec:samping}
Unlike fixed-point following schemes, our method performs target-centric adaptive trajectory sampling, highlighting the multi-objective evaluation over a dense set of candidate points (as shown in Fig.~\ref{fig:sampling}), expanding the feasible solution space and improving flexibility in dynamic, crowded environments. The process consists of three steps: 1) sampling candidate following points, 2) evaluating them using a multi-objective cost function, and 3) estimating the robot’s person-following trajectory based on the optimal point and predicted target trajectory.
\subsubsection{\textbf{Candidate Point Sampling}}
Here, we employ a low-discrepancy point-sampling strategy (Sobol sampling~\cite{sobol1967distribution}) to generate candidate points along the target’s motion direction. These points uniformly populate the sampling space, enabling a comprehensive evaluation of potential following positions. The sampling space is defined as a backward-facing semi-annular region aligned with the target’s motion direction. Its inner and outer radii correspond to the target’s personal and social zones, respectively, thereby satisfying proxemic constraints~\cite{hall1963system} and promoting comfortable, socially compliant following behavior. Specifically, given two consecutive target positions
$\mathbf{p}^\mathrm{Tar}_t=[x^\mathrm{Tar}_t,y^\mathrm{Tar}_t]^\top$ and $\mathbf{p}^\mathrm{Tar}_{t+1}=[x^\mathrm{Tar}_{t+1},y^\mathrm{Tar}_{t+1}]^\top$, 
the target person's heading is: 
\begin{equation} 
\phi \;=\; \operatorname{atan2}\bigl(y^\mathrm{Tar}_{t+1}-y^\mathrm{Tar}_{t},\,x^\mathrm{Tar}_{t+1}-x^\mathrm{Tar}_{t}\bigr). 
\label{eq:heading} 
\end{equation} 
Accordingly, Sobol’ points $u_i=(u_{i,1},u_{i,2})\in[0,1)^2$ are mapped to a semi-annulus whose inner radius $r_{\min}$ corresponds to the target’s personal zone and outer radius $r_{\max}$ to the social zone, thereby ensuring socially comfortable following. The mapping uses an area-uniform radius and a back-half angle, and we sample 50 points to adequately cover the region while maintaining computational efficiency:
\begin{equation} 
r_i=\sqrt{r_{\min}^2+\bigl(r_{\max}^2-r_{\min}^2\bigr)u_{i,1}}, 
~~ 
\theta_i=\phi+\frac{\pi}{2}+\pi\,u_{i,2}. 
\label{eq:polar} 
\end{equation} 
Thus, each candidate following point can be expressed in Cartesian coordinates as:
\begin{equation}
\mathbf{c}_i =
\begin{bmatrix}
x^\mathrm{Tar}_t + r_i \cos\theta_i \\[3pt]
y^\mathrm{Tar}_t + r_i \sin\theta_i \\[3pt]
\theta_i
\end{bmatrix},
\end{equation}
which serve as candidate following positions for subsequent multi-objective evaluation.
\subsubsection{\textbf{Multi-objective Evaluation}}
Subsequently, the sampled target-centric candidate points $\mathbf{c}_i$ are processed by first filtering out those occupied by pedestrians using a local map. The remaining candidates are then assessed through a multi-objective cost function:
\begin{equation}
\label{eq:cost}
\begin{aligned}
\mathrm{min}~J(\mathbf{c}_i)=\;& w_{\mathrm{occ}}  J_{\mathrm{occ}}(\mathbf{c}_i)
      + w_{\mathrm{prox}}  J_{\mathrm{prox}}(\mathbf{c}_i)
      + w_{\mathrm{dist}} J_{\mathrm{dist}}(\mathbf{c}_i) \\
      &+ w_{\mathrm{trav}} J_{\mathrm{trav}}(\mathbf{c}_i)
      + w_{\mathrm{stick}} J_{\mathrm{stick}}(\mathbf{c}_i).
\end{aligned}
\end{equation}
where these weights are empirically set to \( w_{\mathrm{occ}} = 10.0 \), \( w_{\mathrm{prox}} = 1.0 \), \( w_{\mathrm{dist}} = 10.0 \), \( w_{\mathrm{trav}} = 1.0 \), and \( w_{\mathrm{stick}} = 0.5 \).

\textbf{Occlusion Cost.}  
To evaluate whether the target will be occluded when the robot arrives at a candidate position $\mathbf{c}_i$, we assume the robot will reach $\mathbf{c}_i$ after $N$ steps from the current time $t$ using constant velocity. We then construct a virtual camera extrinsic matrix from $\mathbf{c}_i$ and project the predicted positions of the target $\mathbf{p}_{t+N}^T$ and each pedestrian $\mathbf{p}_{t+N}^h$ at time $t+N$ into the panoramic image. The occlusion cost is defined as:
\begin{equation}
J_{\mathrm{occ}}(\mathbf{c}_i)
= \frac{1}{|\mathcal H_N|}
\sum_{h \in \mathcal H_N}
\mathrm{IoU}(\mathcal B_T(\mathbf{p}_{t+N}^T, \mathbf{c}_i),\, \mathcal B_h(\mathbf{p}_{t+N}^h, \mathbf{c}_i)),
\label{eq:occlusion}
\end{equation}
where $\mathcal H_N$ is the set of predicted pedestrians at time $t+N$. $\mathcal B_T(\cdot,\mathbf{c}_i)$ and $\mathcal B_h(\cdot,\mathbf{c}_i)$ are the projected bounding boxes of the target and pedestrian $h$. $\mathrm{IoU}(\cdot,\cdot)$ denotes the intersection-over-union between two projected bounding boxes.

\textbf{Proximity Cost.}  
To ensure the robot avoids intruding into the private zones of surrounding pedestrians, we define the proximity cost:
{\footnotesize
\begin{equation}
\label{eq:proximity}
J_{\mathrm{prox}}(\mathbf{c}_i) =
\begin{cases}
0, & d_{\min}(\mathbf{c}_i) \ge d_{\max}, \\[4pt]
\left(\dfrac{d_{\max} - d_{\min}(\mathbf{c}_i)}{d_{\max} \cdot d_{\min}(\mathbf{c}_i)}\right)^{\!2}, & \varepsilon \le d_{\min}(\mathbf{c}_i) < d_{\max}, \\[6pt]
M, & d_{\min}(\mathbf{c}_i) < \varepsilon,
\end{cases}
\end{equation}
}
where $d_{\min}(\mathbf{c}_i)$ represents the minimum distance between a candidate point $\mathbf{c}_i$ and the predicted pedestrian positions at the corresponding future time step $N$. This distance is calculated as:
\begin{equation}
d_{\min}(\mathbf{c}_i)=\min_{h\in\mathcal H_{\ell(\mathbf{c}_i)}} 
\left\|\,\mathbf{c}_i-\mathbf{p}^h_{\ell(\mathbf{c}_i)}\right\|_2.
\end{equation}
Here, $d_{\max}$ denotes the safety threshold, set to 1.2 m; $\varepsilon$ represents the lower bound to avoid collisions; and $M$ is a large penalty value of 1000, used as a filtering threshold.

\textbf{Distance Cost.}
To maintain an appropriate social distance of target-centric personal zone, we penalize deviations from a desired following distance $d^\ast$, set to $1.5\,\mathrm{m}$ in our experiments. Let $\mathbf{c}^p_i=[c^x_i, c^y_i]^\top$ denote the 2D positional component of candidate $\mathbf{c}i$. The distance cost is defined as:
\begin{equation}
\label{eq:dis}
J_{\mathrm{dist}}(\mathbf{c}_i)
= \left(\,\|\mathbf{c}^p_i - \mathbf{p}^\mathrm{Tar}_t\|_2 - d^\ast\,\right)^2.
\end{equation}

\textbf{Travel Cost.}
The travel cost measures the Euclidean distance between a candidate point and the robot’s current position  $\mathbf{p}^\mathrm{R}_t=[x^\mathrm{R}_t,y^\mathrm{R}_t]^\top$, promoting navigational efficiency by favoring closer goals:
\begin{equation}
\label{eq:travel}
J_{\mathrm{trav}}(\mathbf{c}_i) = \left|\mathbf{c}^p_i - \mathbf{p}^{\mathrm{R}}_t\right|_2.
\end{equation}

\textbf{Stickiness Cost.}
To promote motion smoothness and discourage abrupt changes in the selected point, the stickiness cost is defined as:
\begin{equation}
\label{eq:stick}
J_{\mathrm{stick}}(\mathbf{c}_i)=
\begin{cases}
\left\|\mathbf{c}_i-\mathbf{c}_{\text{prev}}\right\|_2, & \text{if } \mathbf{c}_{\text{prev}} \text{ exists},\\
0, & \text{otherwise.}
\end{cases}
\end{equation}

Through the above sampling and evaluation process, we select the optimal following point that minimizes the multi-objective cost function (Eq.~\ref{eq:cost}).

\subsubsection{\textbf{Person-Following Trajectory Estimation}}  
Upon determining the optimal following point, instead of merely tracking the point as in other methods, we calculate the relative distance \(d\) and orientation \(\theta\) with respect to the target. Based on this, a goal trajectory $G$ is estimated to align with the predicted target trajectory, enabling the lower MPPI local planner to achieve smoother motion.

\subsection{Prediction-Aware MPPI Controller}\label{sec:mppi}
Given the estimated person-following trajectory, we employ a prediction-aware MPPI planner for simultaneous tracking and collision avoidance.
We adopt MPPI \cite{williams2017information} because this sampling-based MPC ensures faster computation and avoids solver failures, making it well-suited for real-time navigation in dynamic environments. For completeness, we briefly introduce our prediction-aware MPPI in the following section.

For the robot dynamic model, we assume a differential model defined as:
\begin{equation}
\mathbf{x}_{k+1} = f(\mathbf{x}_k,\mathbf{u}_k) =
\begin{bmatrix}
x^{\mathrm{R}}_k + v_k \cos\theta_k\,\Delta t \\
y^\mathrm{R}_k + v_k \sin\theta_k\,\Delta t \\
\theta_k + \omega_k\,\Delta t
\end{bmatrix},
\end{equation}
where $\mathbf{x}_k=[x^\mathrm{R}_k,y^\mathrm{R}_k,\theta_k]^\top$ denotes the robot state, serving as the foundation for forward simulation of the MPPI planner. As for the cost function, we introduce three cost terms:

\begin{equation}
\small
\mathrm{min}~J(k) = \sum_{k=0}^{K-1} \big(w_{\mathrm{goal}} J_{\mathrm{goal}}(k) + w_{\mathrm{enc}} J_{\mathrm{enc}}(k) + w_u J_u(k)\big),
\end{equation}
\begin{equation}
\text{s.t.} \quad u_{\mathrm{min}} \leq \mathbf{u}_k \leq u_{\mathrm{max}}, \nonumber
\end{equation}
where $k=0,\ldots,K-1$ indexes the horizon, 
$w$ are the cost weights, 
and $J$ are the cost functions, defined as follows. 

\textbf{Goal Tracking Cost.}  
To ensure smooth and continuous motion, the robot follows the estimated person-following trajectory aligned with the MPPI horizon. At each step, the robot's position $\mathbf{p}^\mathrm{R}_k=[x^\mathrm{R}_k,y^\mathrm{R}_k]^\top$ is encouraged to follow the corresponding goal point $\mathbf{g}_k \in G $:
\begin{equation}
J_{\mathrm{goal}}(k) = \|\, \mathbf{p}^\mathrm{R}_k - \mathbf{g}_k \,\|_2^2.
\end{equation}

\textbf{Encroachment Cost.}  
To avoid entering pedestrians' private zones, we penalize candidate states that fall within a safety margin of $d_{\mathrm{safe}} = 1.2$ m. This margin accounts for both the robot's and pedestrians' physical dimensions around their predicted positions:
\begin{equation}
J_{\mathrm{enc}}(k) =
\sum_{j=1}^{N_h}
\max\!\Big\{0,\; d_{\mathrm{safe}}^2 - \|\, \mathbf{p}^\mathrm{R}_k - \mathbf{p}^{h_j}_k \,\|_2^2 \Big\}.
\end{equation}

\textbf{Control Effort Cost.}  
To further ensure smooth motion and avoid abrupt accelerations, we penalize large control inputs $\mathbf{u}_k$:
\begin{equation}
J_{\mathbf{u}}(k) = \|\, \mathbf{u}_k \,\|_2,
\qquad \mathbf{u}_k=\begin{bmatrix}v_k\\ \omega_k\end{bmatrix},
\end{equation}
where $v_k$ and $\omega_k$ are the linear and angular velocities at time step $k$. 

These cost functions further refine the person-following trajectory by integrating dynamic constraints and prediction-aware collision avoidance, enabling the robot to proactively follow the target person in dynamic and crowded environments.
\section{EXPERIMENTS}
To evaluate our Adap-RPF, we conduct experiments on a public benchmark. This section first introduces the benchmark (\ref{sec:benchmark}), baselines (\ref{sec:baseline}), and implementation details (\ref{sec:implement}). We then compare our approach against fixed point, multiple predefined points and optimization-based baselines (\ref{sec:metrics}), followed by ablation studies analyzing the role of human trajectory prediction in explicit dynamic point selection and local planning (\ref{sec:ablation}). Finally, we demonstrate the system's real-world effectiveness in dynamic crowded environments with a mobile robot, as shown in Fig.~\ref{fig:figure1}.
\newcommand{\best}[1]{\textbf{#1}}
\newcommand{\secondbest}[1]{\underline{#1}}
\begin{table*}[t]
\centering
        \caption{Comparison of task success rates (\%) across four benchmark scenarios with varying pedestrian numbers ($5$–$30$). The task success rate represents the proportion of successfully completed following tasks under different pedestrian numbers in the benchmark scenarios. \textbf{Bold} indicates the best result, while \underline{underlined} marks the second-best.}
\label{tab:detectability}
\setlength{\tabcolsep}{2.5pt} 
\renewcommand{\arraystretch}{1.25}
\large 
\resizebox{\linewidth}{!}{
\begin{tabular}{l*{6}{c}*{6}{c}*{6}{c}*{6}{c}}
\toprule
\multirow{2}{*}{\textbf{Method}}
 & \multicolumn{6}{c}{\textbf{Circular}}
 & \multicolumn{6}{c}{\textbf{Crowd}}
 & \multicolumn{6}{c}{\textbf{Parallel}}
 & \multicolumn{6}{c}{\textbf{Perpendicular}} \\
\cmidrule(lr){2-7}\cmidrule(lr){8-13}\cmidrule(lr){14-19}\cmidrule(lr){20-25}
 & \textbf{5} & \textbf{10} & \textbf{15} & \textbf{20} & \textbf{25} & \textbf{30}
 & \textbf{5} & \textbf{10} & \textbf{15} & \textbf{20} & \textbf{25} & \textbf{30}
 & \textbf{5} & \textbf{10} & \textbf{15} & \textbf{20} & \textbf{25} & \textbf{30}
 & \textbf{5} & \textbf{10} & \textbf{15} & \textbf{20} & \textbf{25} & \textbf{30} \\
\midrule
Wang's \cite{wang2024continuous}
 & \multicolumn{1}{|c}{43.75} & 16.25 & 6.25 & 2.50 & 3.75 & 11.25
 & \multicolumn{1}{|c}{\secondbest{85.00}} & 37.50 & 30.00 & 13.75 & 0.00 & 0.00
 & \multicolumn{1}{|c}{\secondbest{95.00}} & \secondbest{95.00} & \secondbest{90.00} & 75.00 & 70.00 & 40.00
 & \multicolumn{1}{|c}{\secondbest{83.75}} & 38.75 & 17.50 & 10.00 & 3.75 & 0.00 \\
Vu's \cite{vu2025autonomous}
 & \multicolumn{1}{|c}{67.50} & 50.00 & 21.25 & 23.75 & \secondbest{23.75} & \secondbest{20.00}
 & \multicolumn{1}{|c}{78.75} & 57.50 & 25.00 & 13.75 & 16.25 & 2.50
 & \multicolumn{1}{|c}{86.25} & 73.75 & 61.25 & 62.50 & 35.00 & 22.50
 & \multicolumn{1}{|c}{73.75} & 32.50 & 22.50 & 12.50 & 8.75 & 10.00 \\
RDA-Traj \cite{ye2025bench}
 & \multicolumn{1}{|c}{\secondbest{77.50}} & \secondbest{57.50} & \secondbest{35.00} & \best{35.00} & 22.50 & 12.50
 & \multicolumn{1}{|c}{82.50} & \best{82.50} & \secondbest{57.50} & \best{55.50} & \best{35.00} & \best{35.00}
 & \multicolumn{1}{|c}{\secondbest{95.00}} & \secondbest{95.00} & 82.50 & \secondbest{85.00} & \secondbest{72.50} & \secondbest{77.50}
 & \multicolumn{1}{|c}{82.50} & \secondbest{72.50} & \secondbest{47.50} & \secondbest{42.50} & \best{32.50} & \best{25.00} \\
Ours
 & \multicolumn{1}{|c}{\best{83.62}} & \best{64.34} & \best{49.45} & \secondbest{33.68} & \best{47.39} & \best{37.43}
 & \multicolumn{1}{|c}{\best{93.75}} & \secondbest{72.50} & \best{61.25} & \secondbest{42.50} & \secondbest{32.50} & \secondbest{12.50}
 &\multicolumn{1}{|c}{\best{100.00}} & \best{98.75} & \best{98.75} & \best{90.00} & \best{92.50} & \best{85.00}
 & \multicolumn{1}{|c}{\best{90.00}} & \best{78.75} & \best{65.00} & \best{56.25} & \secondbest{31.25} & \secondbest{17.50} \\
\bottomrule
\end{tabular}
}
\label{table:fixed}
\end{table*}
\vspace{0.15\baselineskip}
\subsection{Benchmark}\label{sec:benchmark}
We use a public benchmark \cite{ye2025bench} to evaluate RPF performance under dynamic environments quantitatively, which provides four types of realistic dynamic-crowd scenarios: perpendicular, parallel, circular, and random crowd.
These scenarios are primarily distinguished by pedestrian motion patterns. In the perpendicular setting, pedestrians cross orthogonally to the target’s path, simulating intersections. In the parallel case, they move in the same direction as the target, resembling corridors or sidewalks. In the circular scenario, pedestrians traverse along straight lines connecting opposite points on a circle, mimicking diagonal crossing in plazas or open squares. In the random crowd case, their movements are randomized and disordered, reflecting the chaotic flow in densely populated environments such as shopping malls or train stations. 

\begin{figure}[ht]
\centering
\includegraphics[width=0.4\textwidth]{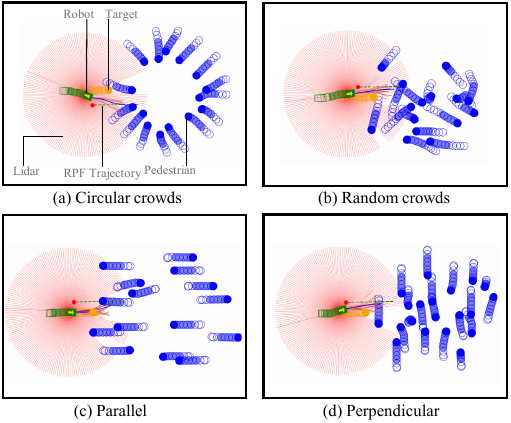}
\caption{\textbf{Representative dynamic-crowd scenarios in the public RPF benchmark:} (a) circular crowds, (b) random crowds, (c) parallel and (d) perpendicular. The Lidar emits red beam lines, while the green dashed line represents the RPF trajectory. Blue circles indicate moving pedestrians, the green rectangle denotes the robot, and the yellow circle marks the target being followed.}
\label{fig:benchmark}
\vspace*{-0.1in}
\end{figure}

In this benchmark, each pedestrian is modeled as a circle with a radius of 0.3 meters. Their interaction behaviors are simulated based on ORCA \cite{van2011reciprocal}. This benchmark presents diverse and challenging conditions, including dynamic occlusions, unpredictable pedestrian trajectories, and frequent crossings. These challenges make it difficult to maintain target visibility, providing a solid evaluation for RPF planning methods. Representative examples are shown in Fig.~\ref{fig:benchmark}.
\begin{figure}[t]
    \centering
\includegraphics[width=0.49\textwidth]{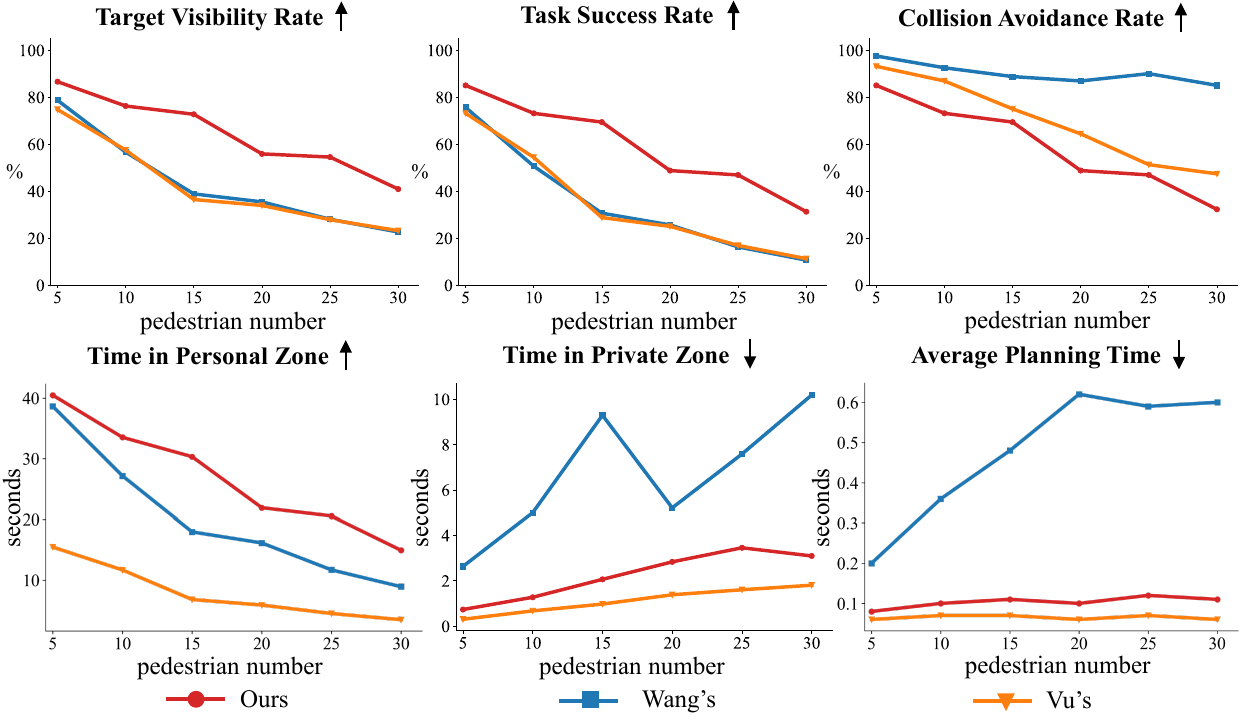}
    \caption{
    \textbf{Comparison of our method with optimization-based and multiple predefine points baselines across six metrics on the public benchmark.} The horizontal axis represents pedestrian number($5$–$30$). Each method is tested across four scenarios: circular crowds, random crowds, parallel, and perpendicular. Results are averaged over 20 random instances per scenario and then across all scenarios.
    }
    \label{fig:comparision}
\end{figure}
\vspace{0.15\baselineskip}
\subsection{Baselines}\label{sec:baseline}
To evaluate the performance of target visibility and following performance on the benchmark, we compare our method with several baselines categorized as follows:
\noindent \subsubsection{Multiple Predefined Following Points}  
\textbf{Vu's Method} \cite{vu2025autonomous} adopts a similar frame-sample-evaluate-plan approach to ours but leverages current scene observations for following points selection and local planning. 

\noindent \subsubsection{Optimization-based}  
\textbf{Wang's Method} \cite{wang2024continuous} is the first to propose maintaining target visibility in RPF by formulating an optimization function within the NMPC framework, incorporating multiple constraints.  

\noindent \subsubsection{Fixed Following Point}  
\textbf{RDA Planner Trajectory Following (RDA-Traj)} achieves SOTA performance in the public benchmark \cite{ye2025bench}. It uses a Kalman-filter-based prediction model to estimate the future trajectory of the target. A person-following trajectory is then extrapolated based on the predicted target trajectory and a fixed following position. The RDA planner \cite{han2023rda} is then used to track this trajectory.

Notably, to ensure a fair comparison of proactive planning capabilities, no person-search techniques \cite{ye2025rpf} are employed; thus, the robot does not attempt to re-find the target once it becomes occluded.

\vspace{0.15\baselineskip}
\subsection{Implementation Details}\label{sec:implement}
We first compute the time required for the target to traverse each dynamic environment without robot following. This duration is scaled by a factor of three, and the maximum time across all environments is used as the evaluation horizon for all algorithms. To ensure diverse and comprehensive testing, we randomly generate 50 scenarios per environment by sampling pedestrian start and end positions within the target’s traversed region. Each scenario includes 5, 10, 15, 20, 25, and 30 pedestrians. All evaluations are performed on a computer with an Intel(R) Core(TM) i9-14900K CPU and NVIDIA GeForce RTX 3090 GPU.

For real robot experiments, we use YOLOX \cite{ge2021yolox} for person detection and a deep re-identification model \cite{ye2024person} to track the target. The platform is a ScoutMini differential-drive robot equipped with an Intel NUC 11 (i7-1165G7 CPU @ 2.80GHz, NVIDIA RTX 2060). A Livox Mid-360 LiDAR (360$^{\circ} \times$59$^{\circ}$ FOV, 10 Hz) and a Ricoh Theta panoramic camera (1280$\times$720, 15 Hz) are mounted on the robot (Fig.~\ref{fig:figure1}). All planners and controllers are configured identically to the simulation setup for consistency.
\begin{table*}[htbp]
\caption{\textbf{Ablation study on TVR and TSR.} 
TVR is the proportion of steps where the target remains visible, and TSR is the percentage of trials where the target is visible at the final step. Both metrics (\%) are evaluated under varying pedestrian counts ($5$ to $30$) to assess the impact of adaptive sampling (Adap.) and trajectory prediction (Traj. Pred.).}

\centering
\setlength{\tabcolsep}{4pt}
\renewcommand{\arraystretch}{1.4}
\footnotesize
\begin{tabular}{cc*{6}{c}*{6}{c}}
\toprule
\multicolumn{2}{c}{\textbf{Ablation}} 
 & \multicolumn{6}{c}{\textbf{TVR / \%}}
 & \multicolumn{6}{c}{\textbf{TSR / \%}} \\
\cmidrule(lr){1-2}\cmidrule(lr){3-8}\cmidrule(lr){9-14}
\textbf{Adaptive Trajctroy Sampling} & \textbf{MPPI Controller}
 & \textbf{5} & \textbf{10} & \textbf{15} & \textbf{20} & \textbf{25} & \textbf{30}
 & \textbf{5} & \textbf{10} & \textbf{15} & \textbf{20} & \textbf{25} & \textbf{30} \\
\midrule
w/o Adap.      & w/ Traj. Pred. 
           & \multicolumn{1}{|c}{86.15} & 77.18 & 61.92 & 52.98 & 38.89 & 41.03
           & \multicolumn{1}{|c}{85.00} & 75.94 & 59.06 & 48.44 & 30.32 & 31.56 \\
w/o Traj. Pred.  & w/o Traj. Pred.     
           & \multicolumn{1}{|c}{62.56} & 46.33 & 29.61 & 25.41 & 21.86 & 19.56
           & \multicolumn{1}{|c}{58.44} & 38.44 & 19.38 & 14.06 & 8.12  & 5.94 \\
w/o Traj. Pred. & w/ Traj. Pred.   
           & \multicolumn{1}{|c}{81.52} & 74.50 & 59.36 & 50.01 & 40.64 & 36.06
           & \multicolumn{1}{|c}{80.62} & 74.06 & 55.94 & 44.68 & 33.44 & 24.06 \\
w/ Traj. Pred.  & w/o Traj. Pred.  
           & \multicolumn{1}{|c}{78.32} & 61.90 & 47.22 & 40.17 & 34.02 & 32.78
           & \multicolumn{1}{|c}{77.50} & 58.44 & 42.50 & 31.88 & 25.31 & 21.25 \\
w/ Traj. Pred.  & w/ Traj. Pred.   
           & \multicolumn{1}{|c}{\textbf{91.96}} & \textbf{79.68} & \textbf{70.40} & \textbf{58.76} & \textbf{55.68} & \textbf{44.26}
           & \multicolumn{1}{|c}{\textbf{91.25}} & \textbf{77.18} & \textbf{65.94} & \textbf{51.88} & \textbf{47.82} & \textbf{34.38} \\
\bottomrule
\end{tabular}%
\label{tab:ablation}
\end{table*}

\subsection{Experiment Results}\label{sec:experiment}
\vspace{0.15\baselineskip}
\subsubsection{Evaluation Metrics}\label{sec:metrics}
We use six common metrics for evaluation. (1) Target visibility rate (TVR) \cite{wang2024continuous}, the proportion of steps where the target remains visible; (2) Collision avoidance rate (CVR), the proportion of scenarios where the robot avoids any collision; (3) Task success rate (TSR), the percentage of trials where the target is still visible at the final step; (4) Time in personal zone (TPZ), the number of steps the robot stays within the target’s personal zone (0.45–1.2 m) \cite{hall1963system}; (5) Time in private zone (TPrZ), the number of steps the robot intrudes into any pedestrian’s intimate zone (0–0.45 m) \cite{hall1963system}; and (6) Average planning time (APT), the average planning runtime per scenario.

\begin{figure}[ht]
\centering
\includegraphics[width=0.4\textwidth]{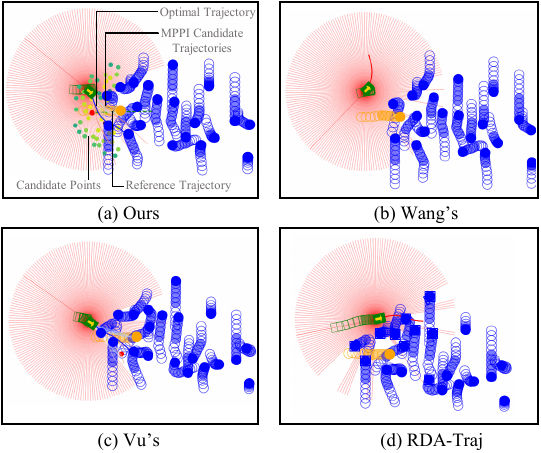}
\caption{\textbf{Comparison of visualization results in the perpendicular scenario with 20 pedestrians.} 
(a) Ours: adaptively switches to the optimal red point to avoid dynamic occlusion; 
(b) Wang's: produces incorrect solutions; 
(c) Vu's: collides with pedestrians; 
(d) RDA-Traj: uses a fixed point, directly runs into the crowd, and loses the target.}
\label{fig:visualize}
\vspace*{-0.1in}
\end{figure}
\vspace{0.15\baselineskip}
As shown in Fig.~\ref{fig:comparision}, our Adap-RPF method significantly surpasses existing SOTA proactive RPF approaches, including predefined-point and optimization-based strategies, across key metrics: TSR, TVR, and TPZ.
For TSR, Adap-RPF achieves approximately 10\% higher performance than the baselines with five pedestrians. Under the 15-pedestrian setting, it reaches 69.38\%, outperforming Wang’s method by 38.76\% and Vu’s by 40.63\%. Even in high-density scenarios with 25 pedestrians, it remains robust at 46.88\%, exceeding both baselines by over 30\%.
For TVR, Adap-RPF maintains a 10\% lead at five pedestrians and reaches 72.70\% with 15 pedestrians, surpassing Wang by 33.90\% and Vu by 36.28\%. At a density of 25 pedestrians, its TVR stays above 50\%, while both baselines fall below 30\%, yielding a margin of over 26\%.
For TPZ, Adap-RPF achieves 40.47 seconds with five pedestrians— higher than Wang (+1.5 $\mathrm{s}$) and 50\% more than Vu. With 15 pedestrians, it records 30.31 seconds, representing 1.7 and 4.5 times the performance of Wang and Vu, respectively. 
These results demonstrate that Adap-RPF ensures superior tracking, target visibility, and social compliance across varying pedestrian densities. Intuitive visualizations further validate these findings (Fig.~\ref{fig:visualize}). In a perpendicular scenario with 20 pedestrians, Adap-RPF selects optimal trajectories, while Wang’s method generates incorrect solutions. Vu’s method causes collisions, and the fixed-point RDA-Traj approach loses the target in dense crowds. While Adap-RPF demonstrates exceptional performance across key metrics, our APT and TiPrZ lag slightly behind Vu’s method. However, they remain real-time and offer more comprehensive evaluations, including target visibility and social comfort. Although the CRT is lower due to the relatively aggressive nature of our approach, it consistently achieves high TVR and TSR.

As summarized in Table.~\ref{table:fixed}, we compare our Adap-RPF method with Wang's method, Vu's method, and the fixed-point baseline RDA-Traj, averaging TSR across fixed-back and side-following modes in various scenarios with different pedestrian densities. While our method outperforms others in most cases, RDA-Traj occasionally performs better in some crowded and perpendicular scenarios. This advantage comes from the fixed-back strategy, where the robot follows directly behind the target, leveraging the target’s natural obstacle avoidance to create a clear path without active planning. Our method does not currently account for environmental patterns, which limits its performance in such cases. Future work will aim to incorporate environmental patterns into adaptive trajectory sampling for further improvement.

\subsubsection{Ablation Study of Our Adap-RPF  Method}\label{sec:ablation}
To assess the contributions of adaptive trajectory sampling and human trajectory prediction for it and MPPI, we conduct ablation studies using TVR and TSR as evaluation metrics. Ablation~1 replaces adaptive trajectory sampling with a fixed relative following point, and constructs a trajectory by maintaining constant distance and orientation along the predicted target path; MPPI still uses human trajectory prediction. Ablation~2 removes trajectory prediction entirely from both adaptive trajectory sampling and MPPI. Ablation~3 disables prediction in adaptive trajectory sampling but retains it in MPPI. Ablation~4 enables prediction for adaptive trajectory sampling but disables it in MPPI. These variants isolate the effects of adaptive trajectory sampling and the role of human trajectory prediction. Results are shown in Table~\ref{tab:ablation}.

We observe that ablation~1 achieves 86.2\% TVR and 85.0\% TSR, below our method (91.9\% TVR and 91.3\% TSR) under the low-density setting with five pedestrians. As the number of pedestrians increases to 15–20, both metrics drop more sharply, indicating that our method remains effective under frequent occlusions. However, performance declines at 30 pedestrians due to overcrowding, suggesting the need to incorporate crowd flow information. These results emphasize the importance of adaptive trajectory sampling. 

In the five-pedestrian scenario, ablation~2 performs the worst, with TVR and TSR dropping to 62.56\% and 58.44\%, more than 25--30\% below our method. Ablation~3 and ablation~4 perform reasonably well under five pedestrians, with TVR at 81.52\% and 78.32\%, and TSR at 80.62\% and 77.50\%, respectively. However, their performance drops sharply as density increases, with TVR falling below 60\% and TSR below 56\%, which is 10–20\% lower than our method. These results from ablation~2 to ablation~4 highlight that human trajectory prediction is critical for both adaptive trajectory sampling and the MPPI controller. The best performance is achieved when these components are effectively combined.
\subsection{Discussion}
As shown in Fig.~\ref{fig:comparision}, the proposed Adap-RPF outperforms optimization-based and predefined-point baselines in TVR, TSR, and TPZ across varying pedestrian densities. This is achieved by sampling candidate points within the target's social zones and evaluating them using a multi-objective cost function that considers target visibility, human comfort, and motion smoothness. The optimal point generates a smooth, socially compliant trajectory, tracked by a prediction-aware MPPI controller. Ablation results (Table.~\ref{tab:ablation}) further highlight that trajectory prediction is essential for both adaptive sampling and MPPI, playing a key role in maintaining stable visibility and robust following.
Nonetheless, as shown in Table.~\ref{table:fixed}, Adap-RPF does not fully exploit the natural shielding effect of following directly behind a person. This limitation leads to performance degradation in highly crowded scenarios, highlighting an area for improvement.
Future work will explore leveraging the crowd flow information to further improve generalization and robustness.
\section{CONCLUSIONS}
In this paper, we present Adap-RPF, a hierarchical planning framework that proactively integrates multiple objectives—target visibility, collision avoidance, motion smoothness, and human comfort. By densely sampling candidate points within the target-centric social space and evaluating them using predicted trajectories of both the target and surrounding pedestrians, Adap-RPF generates a proactive following trajectory that guides a prediction-aware MPPI planner, enhancing both comfort and safety. A key advantage of our method is its ability to maintain target visibility for as long as possible, even under dynamic occlusions, which is crucial for robust and consistent person following in crowded environments. Compared with baseline methods, Adap-RPF achieves SOTA person-following performance across multiple metrics in diverse dynamic scenarios on public benchmarks and real-world experiments.


\bibliographystyle{IEEEtran}
\balance
\bibliography{IEEEabrv, ref}
\end{document}